# When to Retrain: An Empirical Study of Retraining Policies for Streaming ML Under Concept Drift, Budget, and Latency Constraints

Sawan Dasari
Sawan.dasari@gmail.com

## Abstract

Production machine learning systems degrade under concept drift, yet practitioners have little principled guidance on when to retrain. Retraining is costly, retraining budgets are finite, and a retrained model does not take effect instantly: training and deployment latency leave a stale model serving predictions while the data continues to move. We present a controlled empirical study of three practical model-refresh policies (periodic retraining, error-threshold triggering, and statistical drift-triggered retraining with ADWIN) against a no-retrain baseline, evaluated under a unified system model that makes retraining budgets and training-plus-deployment latency explicit. Across 3,933 experiment runs spanning three drift regimes (abrupt, gradual, recurring), three budget levels, up to five latency levels, three datasets (a synthetic benchmark and two real-world streams: LUFlow network-intrusion detection and LendingClub loan default), and two learning modes (with and without per-sample incremental updates), we find that the single most consequential design decision is not the retraining policy but whether the deployed model learns incrementally. With per-sample incremental updates, and for the linear online learner with immediate labels studied here, no policy differs from the no-retrain baseline by a practically significant margin in any of 54 paired comparisons, even at extreme latency. Without incremental updates, policy choice separates outcomes by 15–55 percentage points of post-drift accuracy, and simple periodic retraining significantly outperforms both reactive policies under abrupt and gradual drift, while reactive policies retain an advantage only under recurring drift. We further document systematic failure modes of reactive policies (pre-drift budget exhaustion by error thresholds; gradual-drift blindness and severe budget under-utilization by ADWIN), along with a latency-budget queueing interaction that silently halves effective retraining budgets. We release the full simulator, dataset pipelines, and all per-run artifacts for reproducibility.

## 1. Introduction

An online fraud model watches its error rate climb five percent over an afternoon. Should the team retrain immediately, wait for stronger evidence, or hold to next week's scheduled refresh? Every production ML team faces this decision, and most resolve it with ad-hoc heuristics: retrain nightly, retrain when a dashboard alarm fires, or retrain when someone notices. The decision is consequential because the underlying data distribution shifts constantly (concept drift), retraining consumes real compute and engineering budget, and, critically, retraining is not instantaneous. Between the moment a retrain is triggered and the moment the new model is serving traffic, the old, stale model continues to make predictions on drifted data.

The academic literature offers only partial guidance here. Drift-detection research focuses largely on detecting that drift occurred rather than on deciding when acting on a detection is worth its cost. Theoretical treatments of cost-aware retraining tend to assume abstract cost models and typically ignore training and deployment latency. Surveys catalogue detectors but rarely compare retraining decisions under a common system model. Meanwhile, production systems commonly default to periodic retraining, in part because there is little empirical evidence about when something smarter is actually better.

## 1.1 The Gap

Three ingredients are almost never studied together: (1) a finite retraining budget, a hard cap on the number of retrains available over a stream; (2) explicit latency, a training window during which the old model keeps serving, followed by a deployment delay; and (3) the learning mode of the deployed model, whether it also adapts incrementally between full retrains. Our study makes all three first-class experimental factors and asks a simple question:

*How do different model-refresh policies trade off accuracy, cost, and latency under concept drift in streaming ML systems with limited retraining budgets?*

## 1.2 Contributions

- C1: A reproducible experimental framework. An open-source streaming simulator with pluggable drift generators (abrupt, gradual, recurring), pluggable retraining policies, an explicit budget accounting and latency/deployment mechanism, and two learning modes (with and without per-sample incremental updates). All 3,933 runs, per-run artifacts, and analysis scripts are public.
- C2: A controlled comparison of practical policies. Periodic, error-threshold, and ADWIN drift-triggered retraining plus a no-retrain baseline, evaluated on a full factorial grid over drift type, budget, latency, and seed, on one synthetic and two real-world streaming tasks (LUFlow network-intrusion detection; LendingClub loan default), with paired statistical significance testing (paired t-tests and Wilcoxon signed-rank with Holm–Bonferroni correction, Cohen's d effect sizes, and a pre-defined practical-significance criterion).
- C3: Actionable findings. (i) For linear online learners with immediate labels, per-sample incremental learning dominates policy choice: with it, no policy beats doing nothing by a practically significant margin. (ii) Without incremental learning, periodic retraining, the simplest possible policy significantly beats both reactive policies under abrupt and gradual drift (by 8.6 and 9.8–16.0 percentage points on synthetic data; up to 27 points on LUFlow), while reactive policies win only under recurring drift. (iii) Reactive policies exhibit systematic pathologies: error-threshold triggers waste most of their budget on pre-drift noise, and ADWIN is blind to gradual drift and strands most of its budget under high latency. (iv) A latency–budget queueing interaction silently halves effective budgets whenever latency exceeds the retraining interval.

## 1.3 Paper Organization

Section 2 formalizes the system model, drift taxonomy, cost model, and metrics. Section 3 reviews related work. Section 4 describes the four policies and their calibration. Section 5 details the simulator, the datasets (including a dataset-suitability screening protocol that rejected three candidate real-world datasets), and the experimental grid. Section 6 presents results, Section 7 discusses implications and limitations, and Section 8 concludes.

## 2. Problem Formulation and System Model

### 2.1 Streaming ML System Model

We model a supervised data stream ($x_t$, $y_t$), t = 1, …, T. At each timestep the currently deployed model fθ predicts $\hat{y}_t$ = fθ($x_t$); the true label $y_t$ then arrives (prequential evaluation) and the per-step error is recorded. A retraining policy observes the error stream and the system state and decides at each step whether to trigger a retrain, subject to a budget of K retrains for the whole stream.

Retraining is not instantaneous. When a retrain is triggered at time t, the new model trains on a buffer of recent data for T_retrain steps and is then deployed after an additional T_deploy steps. Throughout the window [t, t + T_retrain + T_deploy) the old model continues to serve predictions; the new model takes effect only at t + T_retrain + T_deploy. Policies are latency-guarded: no new retrain may be triggered while one is in flight, which, as Section 6.5 shows, creates a queueing constraint that can strand budget. Figure 1 illustrates this timeline.

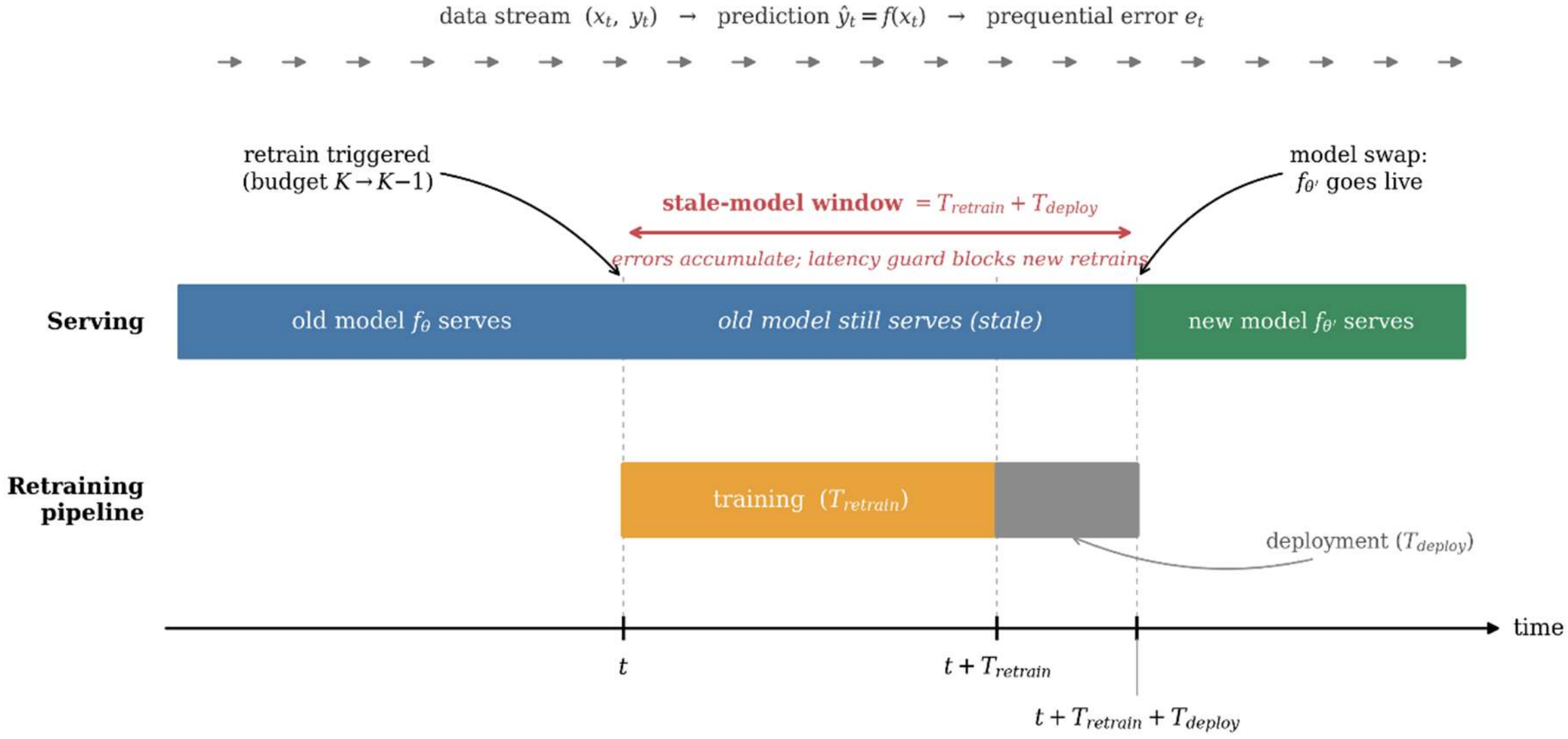

*Figure 1. System-model timeline. A retrain triggered at time t decrements the budget and starts the training pipeline; the old (stale) model continues serving throughout the T_retrain + T_deploy window, and the new model takes effect only at t + T_retrain + T_deploy. The latency guard blocks further triggers while a retrain is in flight.*

### 2.2 Learning Modes

A crucial and usually implicit design choice is what the deployed model does between full retrains. We study both extremes as an explicit experimental factor:

- With incremental updates (partial_fit): the deployed model receives a per-sample gradient update on every observed ($x_t$, $y_t$). Full retrains and incremental learning coexist, as in many online-learning deployments.

- Without incremental updates (static model): the model is frozen between explicit retrains. Only the retraining policy can adapt the model, isolating the pure effect of the policy. This mode reflects the common production pattern of batch-trained models served as immutable artifacts.

## 2.3 Concept Drift Taxonomy

Concept drift is a change over time in P (y | x). We simulate three canonical patterns, chosen to stress different policy failure modes (Figure 2):

- Abrupt drift: a step change in the labeling concept at a single drift point t_d (e.g., a fraud pattern changes overnight).
- Gradual drift: a continuous interpolation from the old to the new concept over a transition window (e.g., user preferences shifting over weeks).
- Recurring drift: the concept alternates periodically between two states after t_d (e.g., seasonal or weekday/weekend effects), with a switch every 1,000 steps in our synthetic configuration.

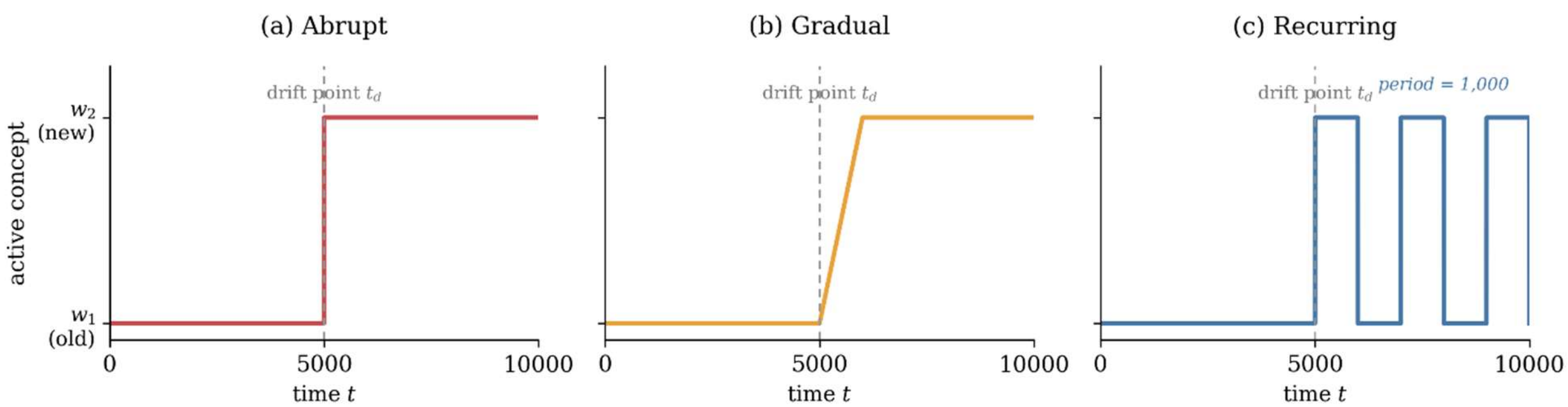

*Figure 2. The three simulated drift patterns (active concept vs. time): (a) abrupt step change at the drift point t_d; (b) gradual transition over a 1,000-step window; (c) recurring concept switches with period 1,000.*

## 2.4 Cost and Latency Model

Each retrain costs one unit against the budget K ∈ {5, 10, 20} for the stream. Latency is modeled as the pair (T_retrain, T_deploy); we study Low (10+1), Medium (100+5), and High (500+20) regimes in the core grid, plus Near-Zero (2+1) and Extreme-High (2000+50) regimes in a dedicated stress phase. For a 10,000-step synthetic stream, K = 5–20 corresponds to a retraining allowance of 0.05–0.2% of the stream. This allowance is deliberately tight, matching real deployments where full retrains are expensive.

## 2.5 Metrics

For each run we record the full error curve and report: overall accuracy; pre-drift accuracy (t < t_d); post-drift accuracy (t ≥ t_d), our primary comparison metric; accuracy drop (post minus pre); total retrains executed; budget utilization (executed / K); and the split of retrains before vs. after the drift point. Post-drift accuracy is primary because it directly measures how well a policy recovers the model after the distribution moves; the before/after retrain split diagnoses why a policy succeeded or failed.

# 3. Related Work

### 3.1 Concept Drift and Drift Detection

Concept drift is a mature research area with comprehensive surveys covering drift categorizations, detection algorithms, and adaptive learning methods [1, 2]. Detection methods fall into three families: sequential-analysis and statistical-process-control detectors such as DDM [3] and EDDM [4]; adaptive-windowing approaches, most prominently ADWIN [5], which maintains a variable-length window and detects change when two sub-windows differ beyond a Hoeffding bound; and distribution-distance methods (Kolmogorov–Smirnov, MMD) applied to features or errors. This literature evaluates detectors on detection delay and false-alarm rates, often implicitly assuming that detection should immediately trigger adaptation. It generally does not model retraining cost, finite budgets, or the latency between triggering and serving a new model, which is the gap this paper addresses.

### 3.2 Retraining Policies and Cost-Aware Adaptation

Periodic retraining is the de facto industrial standard and is a common baseline in applied ML-operations literature [6]. Error-threshold triggering is the classic reactive alternative. Recent work has begun to treat retraining as a constrained-optimization problem, including cost-aware retraining formulations that trade retraining cost against staleness [7, 8]; most recently, the retrain-or-keep decision has been formalized as a sequential decision problem under uncertainty [11]. These works largely abstract away the retraining process itself, treating the model update as taking effect once the decision is made. Our study differs in modeling explicit training and deployment latency (during which the stale model keeps serving) and in empirically characterizing how latency interacts with budgets across drift regimes. We are not aware of prior empirical evaluations that jointly vary retraining budget, training and deployment latency, learning mode, and drift regime as experimental factors while comparing practical refresh policies on both synthetic and real-world streams.

### 3.3 Incremental and Online Learning

Online learning methods adapt per-sample by construction, and the streaming-ML literature [1] treats incremental learners and periodic batch retraining as separate paradigms. Our contribution is to cross the two: we run every policy in both learning modes and show that the presence or absence of per-sample updates changes not just absolute accuracy but the entire policy ranking, a factor that prior policy comparisons have typically not controlled.

### 3.4 ML Systems in Production

Industrial reports document the operational reality of model refresh: retraining pipelines with hours-to-days of end-to-end latency, staleness as a primary cause of performance regressions, and periodic schedules chosen for predictability [6, 9]. Our latency model is a direct abstraction of this pipeline reality, and our findings, particularly that periodic retraining is more robust than reactive triggering when latency is high, give these operational defaults an empirical footing.

## 4. Retraining Policies

We compare three active policies and one baseline. All share the same budget accounting and latency guard; they differ only in the trigger condition.

### 4.1 Periodic Retraining (Baseline Policy)

Retrain every T_period steps, where T_period is derived from the budget so that the schedule spends approximately K retrains over the stream (e.g., $K = 20$ on a 10,000-step stream yields T_period = 500). Periodic retraining is blind to drift: it spends budget uniformly, roughly half before and half after the drift point regardless of drift type. Its virtues are predictability, zero tuning, and immunity to noise-induced false triggers.

### 4.2 Error-Threshold Triggering

Maintain a rolling window of recent prediction errors; trigger a retrain when the windowed mean error exceeds a threshold $\theta$. This is the classic reactive heuristic: spend budget only when performance visibly degrades. Its risks are symmetric, a low threshold fires on noise and exhausts budget before drift, while a high threshold delays reaction and accumulates error.

### 4.3 Drift-Triggered Retraining (ADWIN)

Use the ADWIN change detector [5] on the error stream: maintain an adaptive window, and when two sub-windows differ in mean error beyond a Hoeffding bound at confidence $\delta$, declare drift and trigger a retrain (subject to budget and latency guards). ADWIN's appeal is its principled statistical interpretation $\delta$ directly encodes false-alarm tolerance rather than an empirically tuned threshold.

### 4.4 No-Retrain Baseline

A control policy that never retrains (budget 0). In the incremental mode it isolates the value of per-sample learning alone; in the static mode it provides the accuracy floor that any sensible policy must beat.

### 4.5 Parameter Calibration

Reactive-policy parameters were calibrated per dataset on the abrupt-drift condition using a sweep protocol with three ordered criteria: (1) at most one false-alarm retrain pre-drift, (2) shortest detection delay post-drift, and (3) highest post-drift accuracy as tiebreaker. For the synthetic benchmark, we swept $\theta \in \{0.20, 0.25, 0.27, 0.30, 0.35\}$ (window 200) for the error-threshold policy and $\delta \in \{0.05, 0.01, 0.005, 0.002, 0.001\}$ (window 500, min-samples 300) for ADWIN. The selected values were $\theta = 0.27$, which sits approximately 13% above the empirically measured pre-drift error floor of 23.8% ($\sigma$ = 1.2%) across 81 baseline configurations, and $\delta = 0.002$, which produced zero pre-drift false alarms during calibration while $\delta = 0.001$ occasionally missed abrupt drift entirely. min-samples = 300 skips the warm-up phase in which the freshly initialized learner is still converging. Table 1 summarizes the final parameters. Note that calibration on a small sweep does not guarantee behavior across all seeds: as Section 6.4 shows, thresholds calibrated to be quiet on one seed's noise floor can fire aggressively on another's, which is itself a finding about the fragility of reactive policies.

*Table 1. Calibrated policy parameters (synthetic benchmark). Real-world datasets were calibrated independently with the same protocol.*

| Policy | Parameter | Value |
|---|---|---|
| Periodic | T_period | stream length / K (e.g., 500 at K = 20) |
| Error-threshold | threshold θ | 0.27 |
| Error-threshold | window size | 200 |

| Policy | Parameter | Value |
|---|---|---|
| Drift-triggered (ADWIN) | δ (Hoeffding confidence) | 0.002 |
| Drift-triggered (ADWIN) | window size | 500 |
| Drift-triggered (ADWIN) | min samples (warm-up) | 300 |

# 5. Experimental Framework

## 5.1 Simulator

The simulator implements the streaming loop: at each step, draw ($x_t$, $y_t$) from the drift generator (or real-data stream), predict with the currently deployed model, record the prequential error, let the policy decide, and, in the incremental mode, apply a per-sample partial_fit update. A triggered retrain fits a fresh model on the recent-data buffer and swaps it in only after the full latency window elapses; the latency guard blocks overlapping retrains. The learner is scikit-learn's SGDClassifier with logistic loss throughout, chosen for fast retraining and native incremental-update support. Metrics, per-step traces, and configuration are exported per run as JSON and CSV artifacts.

## 5.2 Synthetic Benchmark

Streams of 10,000 samples with 10 i.i.d. standard-normal features and a binary label generated from a concept weight vector; drift is injected at t = 5,000 by switching (abrupt), interpolating (gradual), or alternating with period 1,000 (recurring) between weight vectors. Synthetic data provides ground-truth drift timing, arbitrary reproducibility, and clean isolation of the experimental factors.

## 5.3 Real-World Datasets and Suitability Screening

Real-world validation of retraining-policy experiments requires more than a public dataset with a timestamp: the dataset must contain a locatable distribution shift that measurably degrades a deployed model, and it must support multiple independent stream constructions (“seeds”) for paired statistics. We formalized this as a three-gate fitness check (task learnability, presence of temporal shift, and shift-induced performance degradation across ≥ 3 stream constructions) and applied it to five candidate datasets. Three failed:

- IEEE-CIS Fraud Detection (590K transactions, 428 features, 3.5% fraud): Kolmogorov–Smirnov tests confirmed strong feature-level shift at the midpoint split (190/428 features shifted at $p < 0.001$), yet task-level degradation was absent or inconsistent across temporal offsets: feature drift did not translate into concept drift. Additionally, the extreme class imbalance made the binary error stream so volatile that ADWIN exhausted its entire budget on pre-drift false alarms in 51 of 54 calibration configurations, a practically important negative result we return to in Section 7.
- Kelmarsh wind-farm SCADA: years with sufficient positive labels were too temporally close to construct multiple independent drift contrasts; six labeling strategies all failed the gates.
- Pump sensor data: the task is trivially separable (99.97%+ accuracy before and after injected drift), leaving no headroom for policies to differ.

Two datasets passed and form our real-world testbed:

- LUFlow network-intrusion detection (Lancaster University): ~21M flow records over 28 daily files with 11 flow-level features (bytes in/out, ports, entropy, inter-packet times), binary benign-vs-malicious labels. We construct 50,000-sample streams with the drift point at t = 25,000 from three pool-pair configurations (class-balance shift, feature drift, and extreme shift), which serve as the seed dimension.
- LendingClub loan default (2007–2018): ~1.35M loans after filtering to terminal outcomes, using 16 origination-time features (34 after one-hot encoding) to avoid post-origination leakage. Streams of 50,000 samples pair year cohorts across the 2012–2016 underwriting-policy changes, giving genuine real-world feature-space drift; three year-pair configurations serve as seeds.

For both real datasets the synthetic drift-injection machinery is additionally applied on top of the pooled streams to realize the abrupt/gradual/recurring factor, and features are standardized with a scaler fit on the pre-drift window.

### 5.4 Experimental Grid

*Table 2. Full experimental scope: 3,933 runs across two learning modes.*

| Phase | Dataset | Grid | Runs |
|---|---|---|---|
| 1. Core (3 seeds) | Synthetic | 3 policies × 3 drifts × 3 budgets × 3 latencies × 3 seeds | 243 |
| 2. Extended (10 seeds) | Synthetic | same grid × 10 seeds | 810 |
| 3. Baseline | Synthetic | no-retrain × 3 drifts × (3 + 10) seeds | 39 |
| 5. Extreme latency | Synthetic | near-zero & extreme-high latency, full grid | 741 |
| 6. LUFlow | LUFlow | 3 pools × 3 drifts × 3 budgets × 3 latencies × 3 policies + baseline | 252 |
| 6. LendingClub | LendingClub | 3 year-pairs × same grid + baseline | 252 |
| 6. Static-model replication | All three | Phases 1/2/4/5 grids without partial_fit | 1,596 |
| Total | | | 3,933 |

### 5.5 Statistical Methodology

All runs use fixed seed sets: 10 seeds (42 … 2021) for synthetic and 3 stream configurations for each real dataset. Every policy is exposed to identical data streams, enabling paired comparisons. For each dataset, learning mode, drift type, and policy pair we compute paired t-tests and Wilcoxon signed-rank tests on per-seed mean post-drift accuracy, apply Holm–Bonferroni correction within each family, and report paired Cohen's d. We pre-define practical significance as $|d| > 0.5$ and $|\Delta \text{ accuracy}| > 2$ percentage points, and treat a comparison as robust only when it is both statistically and practically significant. With n = 3 pairs on the real datasets the Wilcoxon test is unreliable and corrected t-tests are underpowered; for these we emphasize effect sizes and treat significance flags conservatively.

## 6. Results

We organize results around the study's central contrast: the same policies, grids, and datasets evaluated with and without per-sample incremental learning. Unless noted, values are mean post-drift accuracy ± standard deviation across all runs of a cell.

### 6.1 Incremental Learning Nullifies Policy Choice

With per-sample partial_fit enabled, the retraining policy is irrelevant for our linear learner. Across every dataset, all three active policies land within one percentage point of the no-retrain baseline (Table 3, Figure 3). Of 54 paired policy comparisons in this mode, zero meet the practical-significance criterion. Nine reach corrected statistical significance, but every one of them is a sub-1.5-point difference, and in each case the difference slightly favors doing nothing or the cheaper policy (e.g., on LendingClub, no-retrain at 0.715 nominally edges periodic at 0.702, because periodic's full retrains on a small recent buffer briefly discard accumulated knowledge). The pattern holds at every latency level we tested: in the extreme-latency phase, periodic's post-drift accuracy moves from 0.770 at total latency 3 to 0.772 at total latency 2,050, a change within noise across a 683-fold latency increase.

*Table 3. Post-drift accuracy WITH incremental updates. All policies are statistically indistinguishable from never retraining at the practical-significance level.*

| Dataset | Periodic | Error-threshold | Drift-triggered | No-retrain |
|---|---|---|---|---|
| Synthetic (n = 270/30) | 0.771 ± 0.042 | 0.772 ± 0.043 | 0.772 ± 0.042 | 0.774 ± 0.042 |
| LUFlow (n = 81/9) | 0.957 ± 0.020 | 0.962 ± 0.018 | 0.961 ± 0.018 | 0.960 ± 0.017 |
| LendingClub (n = 81/9) | 0.702 ± 0.020 | 0.713 ± 0.021 | 0.704 ± 0.019 | 0.715 ± 0.022 |

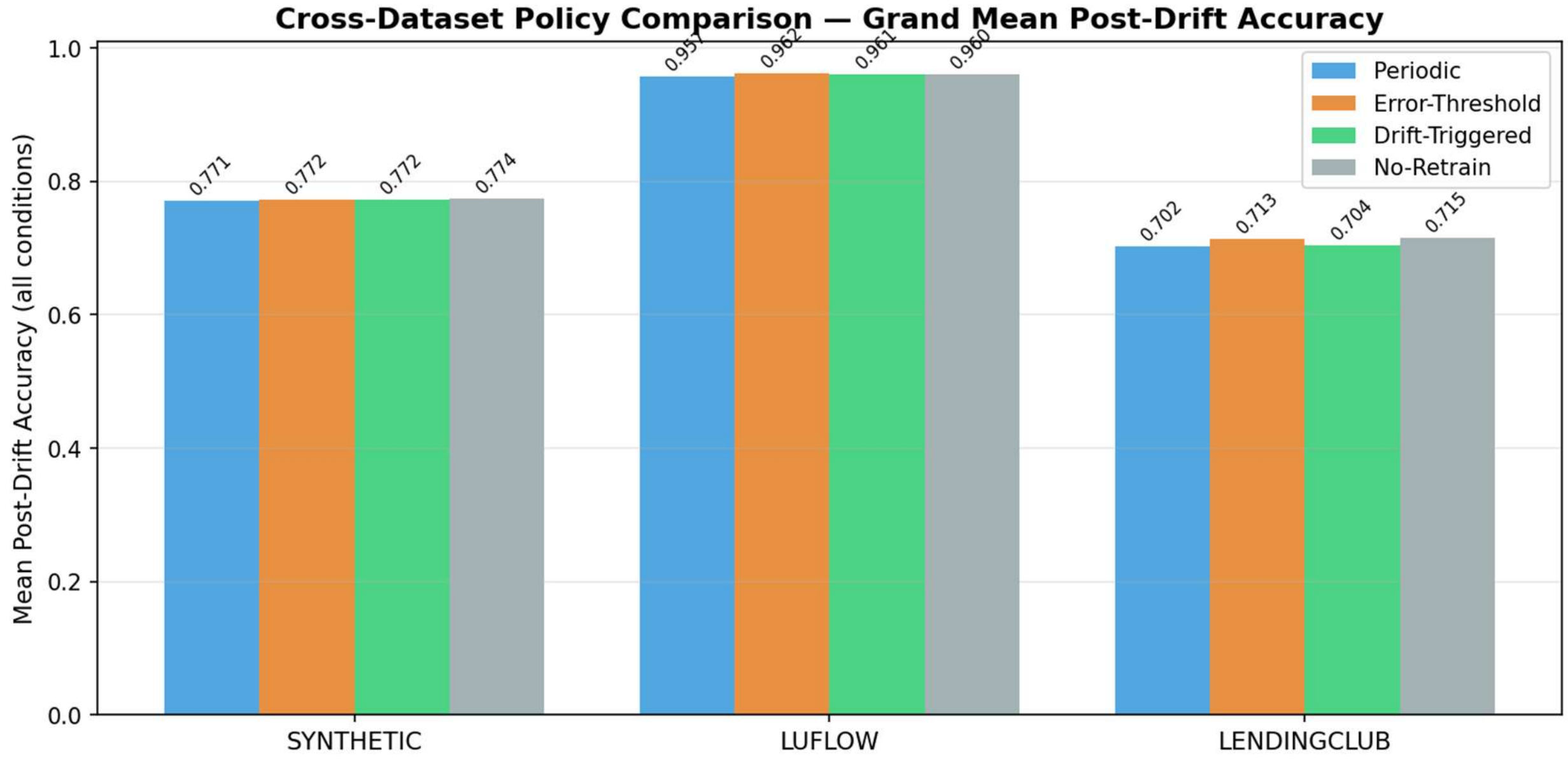

*Figure 3. Cross-dataset post-drift accuracy with incremental updates: policy bars are indistinguishable from the no-retrain baseline on all three datasets.*

The mechanism is straightforward: a per-sample SGD update tracks the moving concept continuously, so by the time any policy triggers a full retrain, the deployed model has already adapted. The practical reading is blunt: if your serving stack can apply cheap incremental updates, that engineering investment buys more than any retraining-trigger sophistication, and an entire class of “when to retrain” machinery becomes unnecessary. We note that this equivalence is strongest when the full retrain and the incremental update share the same hypothesis class and optimizer, as they do for our linear learner and as they may not for deep models fine-tuned or retrained from scratch; Section 7.5 discusses this boundary.

### 6.2 Without Incremental Learning, Policy Choice Separates Dramatically

Freezing the model between retrains transforms the picture (Table 4, Figure 4). The no-retrain floor collapses to 0.505 on synthetic (near chance), 0.287 on LUFlow, and 0.489 on LendingClub. The gap between the best policy and the floor spans 15 to 63 percentage points. In this mode, 44 of 54 paired comparisons are practically significant. Policy choice is now the dominant design decision.

*Table 4. Post-drift accuracy WITHOUT incremental updates. Policies separate by 15–55+ points; periodic retraining leads on every dataset.*

| Dataset | Periodic | Error-threshold | Drift-triggered | No-retrain |
|---|---|---|---|---|
| Synthetic (n = 270/30) | 0.658 ± 0.102 | 0.611 ± 0.118 | 0.581 ± 0.093 | 0.505 ± 0.078 |
| LUFlow (n = 81/9) | 0.844 ± 0.141 | 0.758 ± 0.193 | 0.712 ± 0.211 | 0.287 ± 0.083 |
| LendingClub (n = 81/9) | 0.704 ± 0.025 | 0.667 ± 0.072 | 0.692 ± 0.062 | 0.489 ± 0.043 |

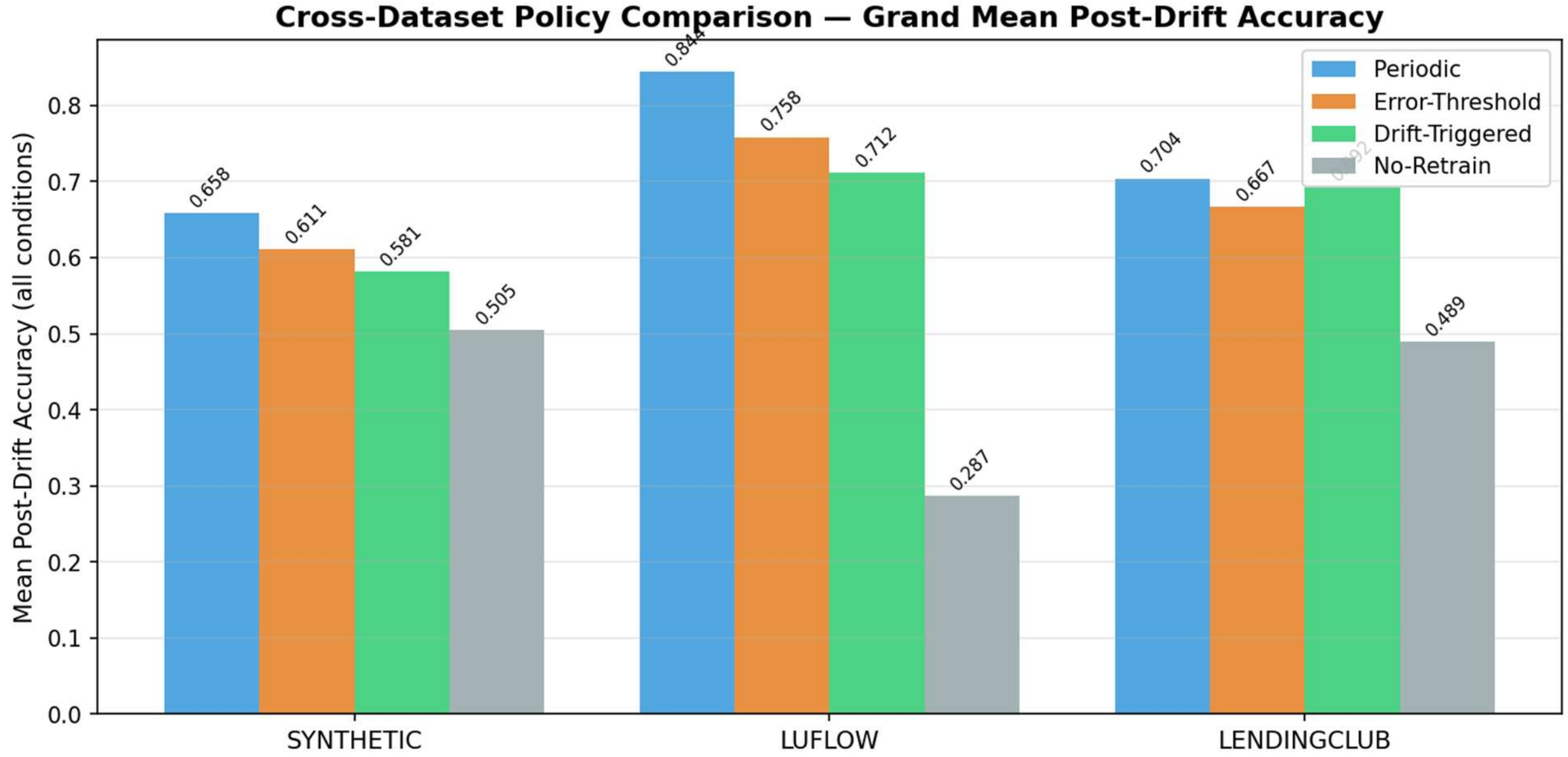

*Figure 4. Cross-dataset post-drift accuracy without incremental updates: large policy separation, with periodic retraining leading and never-retraining collapsing.*

### 6.3 Periodic Retraining Beats Reactive Policies, Except Under Recurring Drift

Breaking the static-model results down by drift type (Table 5, Figure 5) reveals a consistent and counterintuitive ranking. Under abrupt drift, periodic beats error-threshold by 8.6 points (0.703 vs. 0.618; d = 1.55, corrected p = 0.011) and drift-triggered by 9.0 points (d = 1.71, corrected p = 0.006). Under gradual drift the gaps widen: +9.8 points over error-threshold (d = 1.82) and +16.0 points over ADWIN (d = 2.21, corrected p = 0.001). The "smart" reactive policies, the very ones designed to detect and respond to drift, lose to a blind schedule precisely on the drift patterns they were designed for.

*Table 5. Synthetic benchmark without incremental updates: mean post-drift accuracy by policy and drift type (n = 90 runs per cell; 10 per cell for no-retrain).*

| Policy | Abrupt | Gradual | Recurring |
|---|---|---|---|
| Periodic | 0.703 | 0.712 | 0.560 |

| Policy | Abrupt | Gradual | Recurring |
| --- | --- | --- | --- |
| Error-threshold | 0.618 | 0.614 | 0.601 |
| Drift-triggered (ADWIN) | 0.614 | 0.552 | 0.578 |
| No-retrain | 0.489 | 0.501 | 0.525 |

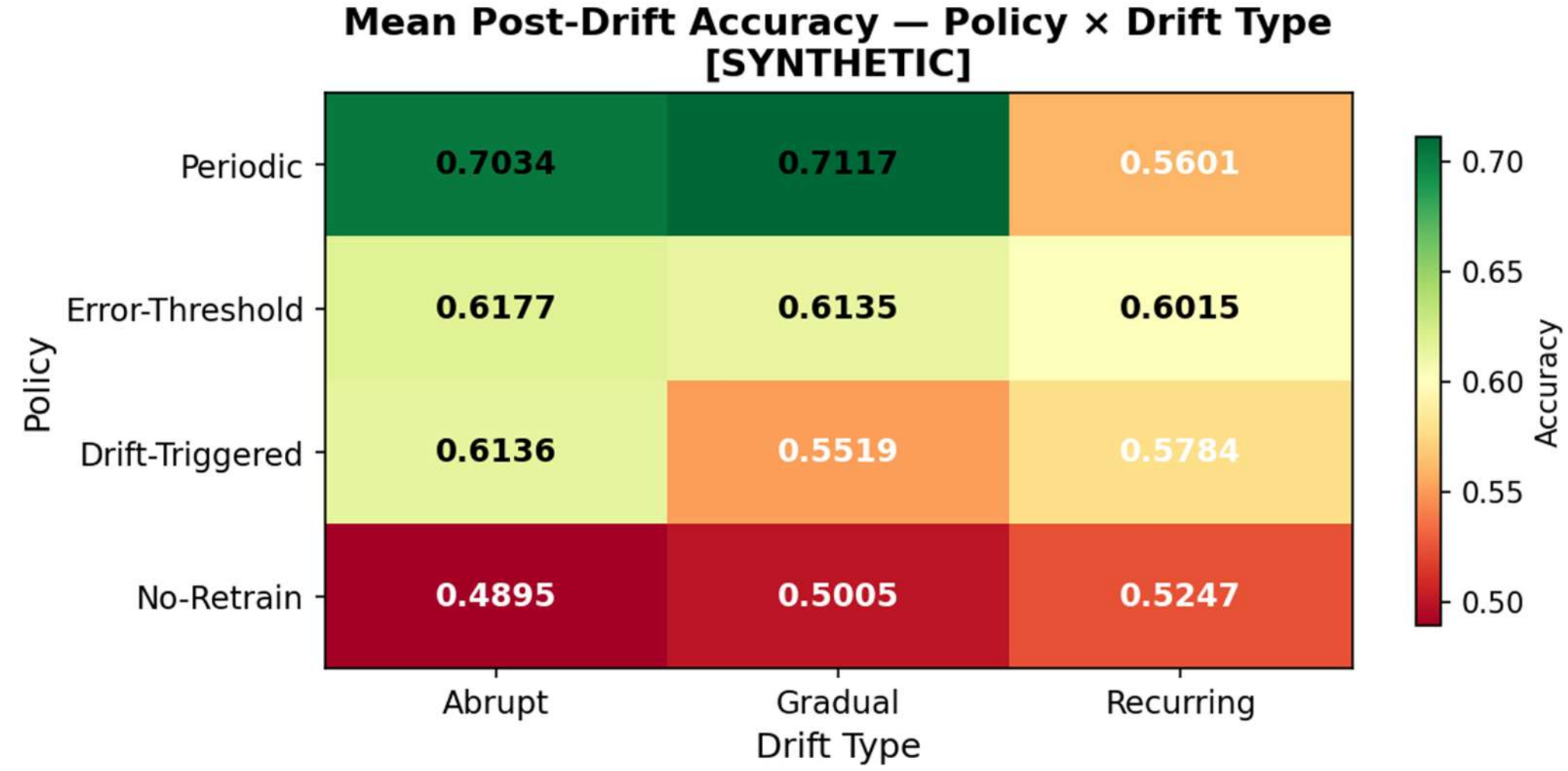

*Figure 5. Policy × drift-type heatmap of post-drift accuracy (synthetic, static model). Periodic dominates abrupt and gradual drift; the ranking inverts under recurring drift.*

Recurring drift inverts the ranking: error-threshold (0.601) and ADWIN (0.578) edge periodic (0.560), though the reversal does not survive multiple-comparison correction (periodic vs. error-threshold: −4.1 points, corrected $p = 0.14$). The mechanism is timing: with the concept switching every 1,000 steps, a fixed schedule aligns with switches only by luck, whereas reactive triggers naturally fire shortly after each switch degrades performance. Recurring drift is also simply the hardest regime for everyone: a budget of $K = 5$ cannot answer five concept switches, and every policy's accuracy is depressed relative to the single-shift regimes.

The real datasets replicate the headline. On LUFlow, periodic reaches 0.884 (abrupt) and 0.901 (gradual) versus 0.780/0.771 for error-threshold and 0.618/0.766 for ADWIN (a 27-point gap on abrupt drift), with effect sizes up to $d = 6.4$, though with $n = 3$ stream configurations the corrected tests are underpowered. On LendingClub, periodic again leads every drift type (0.696–0.711) with the smallest variance of any policy; notably its standard deviation (0.025) is a third of error-threshold's (0.072), meaning the schedule is not just better on average but far more predictable.

## 6.4 Why Reactive Policies Fail: Two Systematic Pathologies

### *6.4.1 Error-threshold triggers exhaust budget before drift*

The error-threshold policy's Achilles heel is pre-drift noise. In the static mode at low latency, it fires on average 11.7 of its retrains before the drift point and zero after: the entire budget is consumed reacting to noise in a phase where retraining cannot help, leaving nothing for the actual drift. Averaged across all static-mode configurations, 66–77% of the error-threshold budget is spent pre-drift (Figure 6), versus a structural ~52% for periodic (which by design splits its schedule around the midpoint drift) and only 14–31% for ADWIN. Threshold behavior is also severely seed-sensitive: in the 3-seed calibration phase, seeds 42 and 123 exhausted all K = 5 retrains before t = 5,000 while seed 456 fired all five after: identical policy, identical parameters, opposite behavior. A fixed threshold encodes an assumption about the noise floor that individual data realizations freely violate.

### *6.4.2 ADWIN is blind to gradual drift and strands its budget*

ADWIN exhibits the mirror-image failure: it is too conservative. In the incremental mode it detects essentially nothing, averaging 1.1 retrains under abrupt drift and 0.96 under gradual (budget utilization 12% and 9%), because per-sample updates keep the error stream too stable to breach the Hoeffding bound; in the 3-seed phase it triggered zero retrains in all 27 gradual-drift configurations. Even in the static mode, where errors visibly climb, gradual drift produces only 30% budget utilization: the slow, continuous error increase never presents the sharp sub-window contrast the statistical test requires. Detection is also seed-dependent: under abrupt drift, some seeds yield zero detections across all budget-latency cells while others fully utilize the budget, depending on whether the drift magnitude clears that realization's noise. The broader lesson: detectors of this kind answer the question "has the error distribution provably changed?", which is neither necessary nor sufficient for answering "is retraining now worth one unit of budget?".

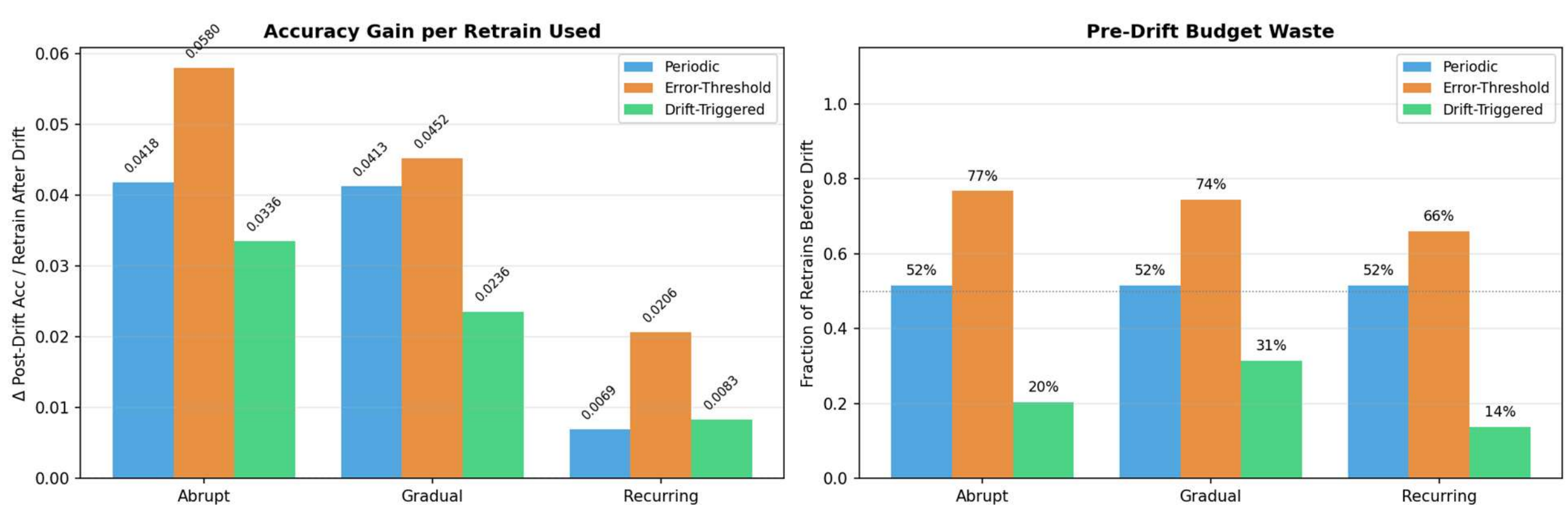

*Figure 6. Budget efficiency (synthetic, static model): accuracy gain per retrain executed and fraction of budget spent before drift. Error-threshold wastes most of its budget pre-drift; ADWIN under-spends.*

## 6.5 Latency Effects: Queueing, Stranded Budget, and a Counterintuitive Reversal

Latency interacts with retraining through a simple queueing mechanism: while a retrain is in flight, no new retrain may start. When the total latency window (520 steps in the High regime) exceeds the periodic interval (500 steps at K = 20), every other scheduled retrain is skipped, the theoretical budget of 20 collapses to an effective ~10, a 50% silent loss. Across the static-mode grid, high latency cuts mean budget utilization to 83% for periodic, 77% for error-threshold, and just 23% for ADWIN, whose late, sparse detections are disproportionately blocked.

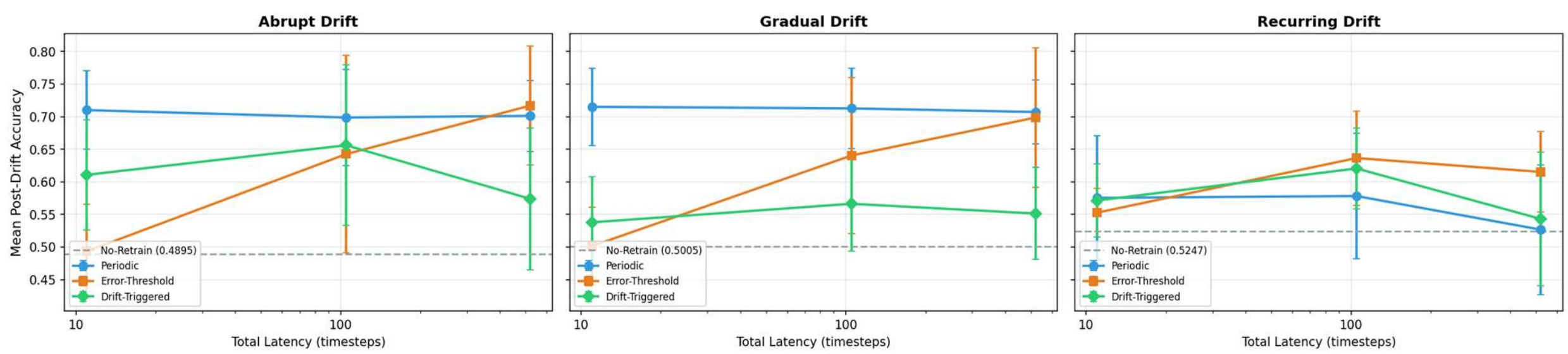

*Figure 7. Latency sensitivity by policy and drift type (synthetic, static model). Periodic degrades only mildly with latency; error-threshold paradoxically improves because queueing throttles its pre-drift false triggers.*

Latency's accuracy impact, however, is smaller and stranger than expected (Figure 7). Periodic loses only 0.8–0.9 points moving from low to high latency under abrupt and gradual drift (4.9 points under recurring, where each 520-step stale window covers half a concept period). The error-threshold policy improves under high latency, by 22.4 points under abrupt drift (0.493 → 0.717), because the latency guard acts as an accidental rate limiter: at low latency the policy freely burns its budget on pre-drift noise (11.7 pre-drift retrains, 0 post-drift), while at high latency queueing throttles pre-drift firing (3.7 pre-drift) and preserves budget for after the drift (4.2 post-drift). High latency inadvertently fixes the policy's false-trigger pathology. This is not an endorsement of slow pipelines; it is evidence that the policy's low-latency behavior was pathological to begin with. In the incremental mode, consistent with Section 6.1, latency has no measurable accuracy effect anywhere in the tested range, including the extreme 2,050-step regime.

### 6.6 Statistical Summary

Table 6 consolidates the significance testing. The asymmetry between modes is the study's central quantitative result: with incremental learning, zero of 54 comparisons are practically significant; without it, 44 of 54 are, and the periodic-over-reactive advantage under abrupt and gradual drift survives Holm–Bonferroni correction on the fully powered 10-seed synthetic grid.

*Table 6. Selected paired comparisons (synthetic, 10 seeds, static model). Δ = periodic minus comparator, post-drift accuracy; p-values Holm–Bonferroni corrected.*

| Comparison | Drift | Δ (pp) | Cohen's d | corrected p | Robust? |
|---|---|---|---|---|---|
| Periodic vs. Error-threshold | Abrupt | +8.6 | 1.55 | 0.011 | Yes |
| Periodic vs. Error-threshold | Gradual | +9.8 | 1.82 | 0.004 | Yes |
| Periodic vs. Error-threshold | Recurring | −4.1 | −0.92 | 0.138 | No |
| Periodic vs. Drift-triggered | Abrupt | +9.0 | 1.71 | 0.006 | Yes |
| Periodic vs. Drift-triggered | Gradual | +16.0 | 2.21 | 0.001 | Yes |
| Periodic vs. No-retrain | Abrupt | +21.4 | 2.41 | <0.001 | Yes |
| Periodic vs. No-retrain | Gradual | +21.1 | 2.72 | <0.001 | Yes |

## 7. Discussion

### 7.1 When Each Policy Wins

*Table 7. Policy suitability matrix derived from the full grid.*

| Deployment scenario | Recommended policy | Rationale | Caveats |
|---|---|---|---|
| Serving stack supports incremental updates | Any / none | Per-sample learning tracks drift continuously; policy choice is practically irrelevant (0/54 significant differences) | Assumes labels arrive with the stream; label delay untested |
| Static model, abrupt or gradual drift | Periodic | +8.6 to +16.0 pp over reactive policies with large effect sizes; lowest variance | Spends ~half the budget before drift by construction |
| Static model, recurring / cyclic drift | Error-threshold | Reactive timing aligns retrains with concept switches (+4.1 pp over periodic) | Advantage not significant after correction; still noise-sensitive |
| High retraining/deployment latency | Periodic | Degrades only ~1 pp; predictable queueing; reactive detections get blocked (ADWIN utilization falls to 23%) | Ensure latency + deploy < interval or effective budget halves |
| Class-imbalanced error streams | Periodic or calibrated threshold | ADWIN's Hoeffding test cannot separate imbalance-driven error volatility from drift (IEEE-CIS: 51/54 configs exhausted budget on false alarms) | Observed during dataset screening, not the full grid |

## 7.2 Surprising Findings

- The biggest lever is not a policy. We began this study to rank retraining policies; the data instead ranked learning modes. In our grid, whether the deployed model updates incrementally separates outcomes more than any policy or hyperparameter choice.
- Simple beats smart on the drift it was built for. Reactive policies exist to handle abrupt and gradual drift, yet those are exactly the regimes where a blind schedule wins by the largest, most significant margins. The reactive policies' losses come not from detection failure per se but from budget mismanagement around detection: firing early (threshold) or not at all (ADWIN).
- High latency can improve a bad policy. The latency guard's rate-limiting accidentally repaired the error-threshold policy's false-trigger pathology, improving abrupt-drift accuracy by 22 points. Constraints can mask policy defects, which also means low-latency tests of a reactive policy may look worse than production under slower pipelines, confusing attribution.
- Statistical rigor in detection does not equal decision quality. ADWIN's principled false-alarm control made it the least effective policy overall: it under-spent budget everywhere, detected gradual drift essentially never, and had its sparse detections blocked by latency.

## 7.3 Implications for Practitioners

First, invest in incremental serving before investing in trigger sophistication: if per-sample (or per-micro-batch) updates are feasible, the entire retraining-policy question largely dissolves. Second, when models must be static between refreshes, periodic retraining should be the default; adopt reactive triggering only when drift is known to be recurring on a timescale your budget can answer. Third, check the queueing condition explicitly: keep T_retrain + T_deploy below your retraining interval, or accept that your effective budget is smaller than your nominal one. Fourth, do not deploy Hoeffding-bound detectors on heavily class-imbalanced error streams without imbalance-aware preprocessing; our screening experience with fraud data suggests their false-alarm behavior can consume the budget before drift arrives.

### 7.4 Dataset Suitability as a First-Class Methodological Concern

Three of five candidate real-world datasets proved unusable for retraining-policy evaluation, each for a different reason: feature drift without task drift (IEEE-CIS), insufficient temporal structure for repeated contrasts (Kelmarsh), and a ceiling-effect task (pump sensors). We suspect this failure rate is typical rather than exceptional, and that some published drift-adaptation results on real data may partly reflect these artifacts. The three-gate fitness protocol of Section 5.3 is cheap to run (a no-retrain baseline sweep plus KS diagnostics) and, we argue, should precede any drift-adaptation evaluation on a real dataset.

### 7.5 Limitations

Our learner is a linear SGD classifier; deep models with longer training times and different fine-tuning dynamics may shift the incremental-vs-static contrast, though the queueing and budget mechanics are model-agnostic. Labels are assumed available immediately after prediction; label delay, ubiquitous in credit and fraud settings, would degrade all reactive policies further and is a priority extension. Retraining cost is a fixed unit rather than a function of data volume or model size. The real-dataset arms use three stream configurations, limiting corrected-test power (we report effect sizes for this reason). Finally, we study pure policies; hybrid schemes (a periodic floor plus drift-triggered top-ups) and budget-aware schedulers are natural future work our framework directly supports.

## 8. Conclusion

We presented a 3,933-run empirical study of practical model-refresh policies for streaming ML under concept drift, explicit retraining budgets, and training-plus-deployment latency, evaluated on synthetic and two real-world streams in both incremental and static learning modes. The central finding reframes the question the study set out to answer: when the deployed model learns incrementally, no retraining policy, including never retraining, differs practically from any other at any latency we could construct, at least for the linear learner and immediate-label setting studied here. When the model is static between refreshes, policy choice moves post-drift accuracy by 15–55+ percentage points, and the simplest policy wins: periodic retraining significantly outperforms error-threshold and ADWIN drift-triggered policies under abrupt and gradual drift, ceding ground only under recurring drift. Reactive policies fail for identifiable, systematic reasons (pre-drift budget exhaustion and statistical conservatism), and latency interacts with budgets through a queueing mechanism every practitioner can check with one inequality. Model refresh is not one-size-fits-all, but its decision tree is shorter than the literature implies: enable incremental learning if you can; schedule periodically if you cannot; and reach for drift detection only when your drift is recurring and your budget can keep up.

All code, per-run artifacts, and analysis pipelines are available at: https://github.com/sa1dasari/Study-of-Drift-Triggered-Retraining-Policies-Under-Budget-and-Latency-Constraints